%% file: main.tex
\documentclass[10pt,twocolumn,letterpaper]{article}

\usepackage[pagenumbers]{cvpr}      
\usepackage{footmisc}
\input{preamble}

\definecolor{cvprblue}{rgb}{0.21,0.49,0.74}
\usepackage[pagebackref,breaklinks,colorlinks,citecolor=cvprblue]{hyperref}

\def\paperID{4769} 
\def\confName{CVPR}
\def\confYear{2024}

\title{A Simple Baseline for Efficient Hand Mesh Reconstruction}

\author{
Zhishan Zhou\textsuperscript{*},
Shihao Zhou\textsuperscript{*},
Zhi Lv ,
Minqiang Zou,
Yao Tang,
Jiajun Liang \textsuperscript{\dag}\\
Jiiov Technology\\
{\tt\small \{ zhishan.zhou, shihao.zhou, zhi.lv, minqiang.zou, yao.tang, jiajun.liang\}@jiiov.com}
\\
\\
\href{http://simplehand.github.io}{http://simplehand.github.io}
}
\begin{document}
\maketitle
\let\thefootnote\relax\footnote{*  Equally contribution. \dag \ Corresponding author.}
\input{sec_simhand/0_abstract}    
\input{sec_simhand/1_intro}
\input{sec_simhand/2_relatedwork}

\input{sec_simhand/3_method}

\input{sec_simhand/4_experiment}
\input{sec_simhand/5_conclution}

{
    \small
    \bibliographystyle{ieeenat_fullname}
    \bibliography{main}
}


\end{document}

%% file: preamble.tex
\usepackage[dvipsnames]{xcolor}


%% file: sec_simhand/0_abstract.tex
\begin{abstract}
Hand mesh reconstruction has attracted considerable attention in recent years, with various approaches and techniques being proposed. Some of these methods incorporate complex components and designs, which, while effective, may complicate the model and hinder efficiency. In this paper, we decompose the mesh decoder into token generator and  mesh regressor. Through extensive ablation experiments, we found that the token generator should select discriminating and representative points, while the mesh regressor needs to upsample sparse keypoints into dense meshes in multiple stages. Given these functionalities, we can achieve high performance with minimal computational resources. Based on this observation, we propose a simple yet effective baseline that outperforms state-of-the-art methods by a large margin, while maintaining real-time efficiency.  Our method outperforms existing solutions, achieving state-of-the-art (SOTA) results across multiple datasets. On the FreiHAND dataset, our approach produced a PA-MPJPE of 5.8mm and a PA-MPVPE of 6.1mm. Similarly, on the DexYCB dataset, we observed a PA-MPJPE of 5.5mm and a PA-MPVPE of 5.5mm. As for performance speed, our method reached up to 33 frames per second (fps) when using HRNet and up to 70 fps when employing FastViT-MA36. Code will be made available.

\end{abstract} 

%% file: sec_simhand/1_intro.tex
 \begin{figure}[ht!]  
    \centering  
    \includegraphics[width=0.5\textwidth]{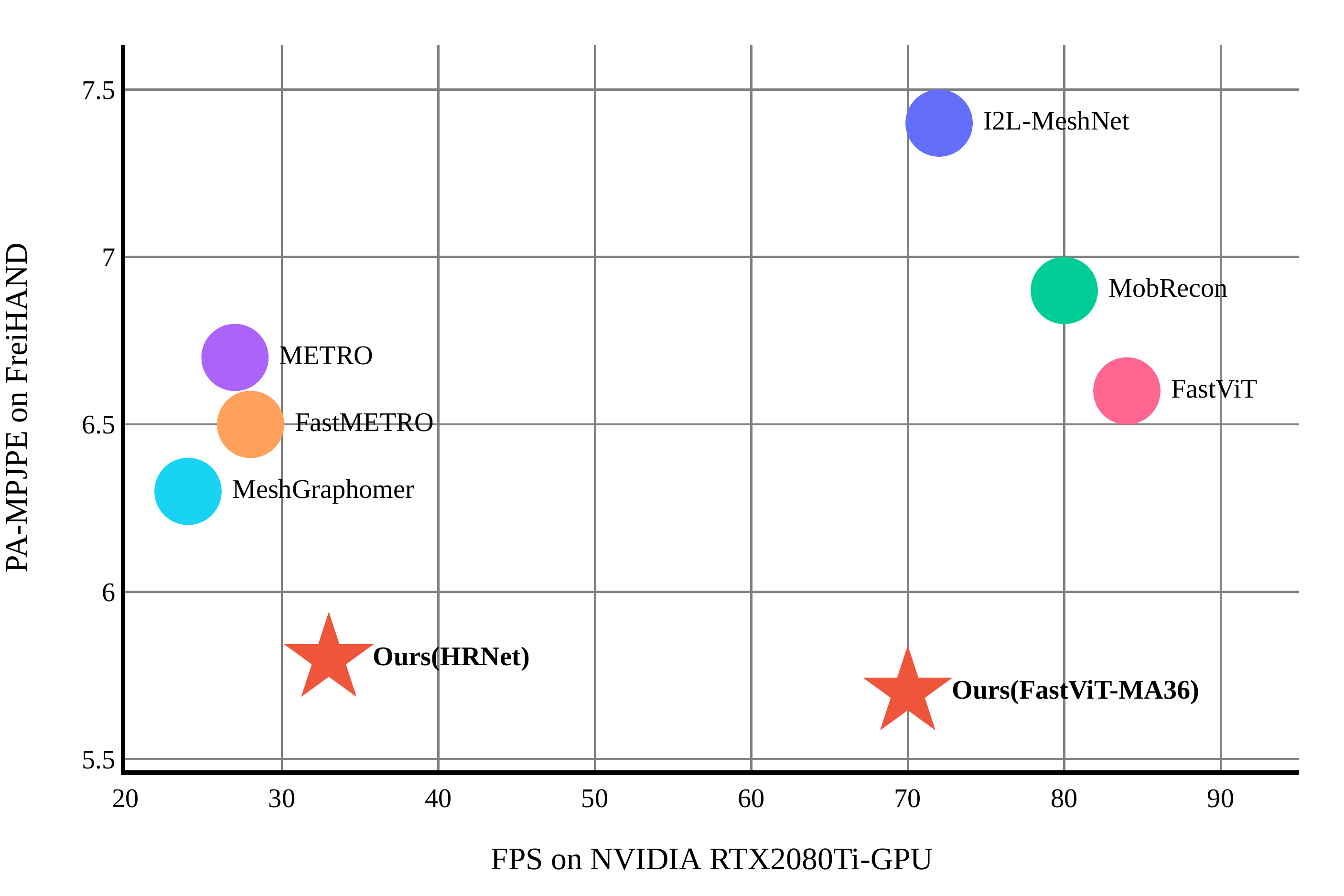}  
    \caption{Trade-off between Accuracy and Inference Speed. Our technique surpasses non-real-time methods($\leq$ 40 fps) in both speed and precision. Compared to real-time methods ($\geq$ 70 fps), it offers a substantial boost in accuracy while preserving comparable speeds. For fair comparison, all speed evaluations were conducted on a 2080ti GPU with a batch size of one.}  
    \label{fig:speed_test}  
\end{figure}

\section{Introduction} 
\label{sec:intro}
The field of hand mesh reconstruction has seen rapid advancements, with various types of mesh decoders being proposed. Despite their commendable performance, these methods often suffer from high system complexity, involving unnecessary components that may hinder efficiency. To facilitate a clear discussion, we decompose the mesh decoder into two primary components: the token generator and the mesh regressor.

The token generator serves a crucial role by integrating prior information with image features to extract task-specific features. For instance, FastMETRO ~\cite{Cho_Youwang_Oh}  employs a strategy to predict weak-perspective camera parameters, which aggregates image features. MobRecon~\cite{Chen_Liu_Dong_Zhang_Ma_Xiong_Zhang_Guo_2022} develops a stacked encoding network to obtain gradually refined encoding features, and applies a technique known as pose pooling to suppress features that are unrelated to joint landmarks. PointHMR~\cite{kim2023sampling} on the other hand, proposes to use features sampled at positions of vertices projected from 3D to 2D spaces as intermediate guidance. These approaches collectively provide informative and discriminating features that enhance the overall performance of the system.

The mesh regressor, the second component, decodes the tokenized features obtained from the token generator into mesh predictions. FastMETRO~\cite{Cho_Youwang_Oh} takes a set of learnable joint tokens and vertex tokens as input and masks self-attentions of non-adjacent vertices according to the topology of the triangle mesh during training. MobRecon~\cite{Chen_Liu_Dong_Zhang_Ma_Xiong_Zhang_Guo_2022} employs a strategy of 2D-to-3D lifting and Pose-to-vertex lifting to gradually approximate meshes. MeshGraphormer~\cite{Lin_Wang_Liu_2021} uses a coarse template mesh for positional encoding and then applies a linear Multi-Layer Perceptron (MLP) to sample the coarse mesh up to the original resolution. These methods aim to alleviate training difficulties due to heterogeneous modalities.

Through investigation on existing methods, we found a interesting phenomenon that, although some methods shares same performance, they differs in specific failure cases. Namely, methods with coarse sampling strategy lack perceptual ability for fine-grained gestures such as pinch. Methods with limited upsample layers struggles in generating reasonable hand shapes. This observation prompts us to question: \textit{\textbf{How different structures make effect on mesh decoder?}} By answering the question, we can streamline the process, eliminating excessive computation and complex components, to complete mesh prediction in a simple and efficient way. To design concise experiments, we start from the simplest structure for the aforementioned two modules, then gradually add and optimize the most commonly used components abstracted from the state-of-the-art (SOTA) methods.

Through extensive ablation experiments, we discovered that the important structure of token generator is to sample discriminating and representative points, while the important structure of mesh generator is to upsample sparse keypoints into dense meshes. For implicitly, in the following paper, we define each of these structure as \textit{\textbf{core structure}}. In the model design process, provided that the core structure's functionality is fulfilled, high performance can be achieved with minimal computational resources.

Based on these observations, we propose a simple baseline that surpasses the SOTA methods by a significant margin and is computationally efficient. Referring to Figure~\ref{fig:speed_test}, our proposed technique delivers state-of-the-art performance on various datasets. On the FreiHAND~\cite{zimmermann2019freihand} dataset, it recorded a PA-MPJPE of 5.8mm and PA-MPVPE of 6.1mm. When tested on the DexYCB~\cite{chao2021dexycb} dataset, these metrics were further refined to a PA-MPJPE of 5.5mm and a PA-MPVPE of 5.5mm. Our method is also advantaged in efficiency, achieving 33 frames per second (fps) on HRNet\cite{wang2020deep} and an impressive 70 fps on FastViT-MA36~\cite{Vasu_Gabriel_Zhu_Tuzel_Ranjan_2023}. 
 Our contributions can be summarized as follows:

\begin{enumerate}
\item We abstract existing methods into token generator and mesh regressor modules, and reveal the core structures of these two modules respectively.
\item Based on these core structures, we developed a streamlined, real-time hand mesh regression module that excels in both efficiency and accuracy.
\item Our method has achieved PA-MPJPE of 5.7mm and PA-MPVPE of 6.0mm on FreiHAND, and achieved SOTA results on multiple datasets, demonstrating its effectiveness and generalizability.
\end{enumerate}

%% file: sec_simhand/2_relatedwork.tex
 \begin{figure*}[ht!]  
    \centering  
    \includegraphics[width=\textwidth]{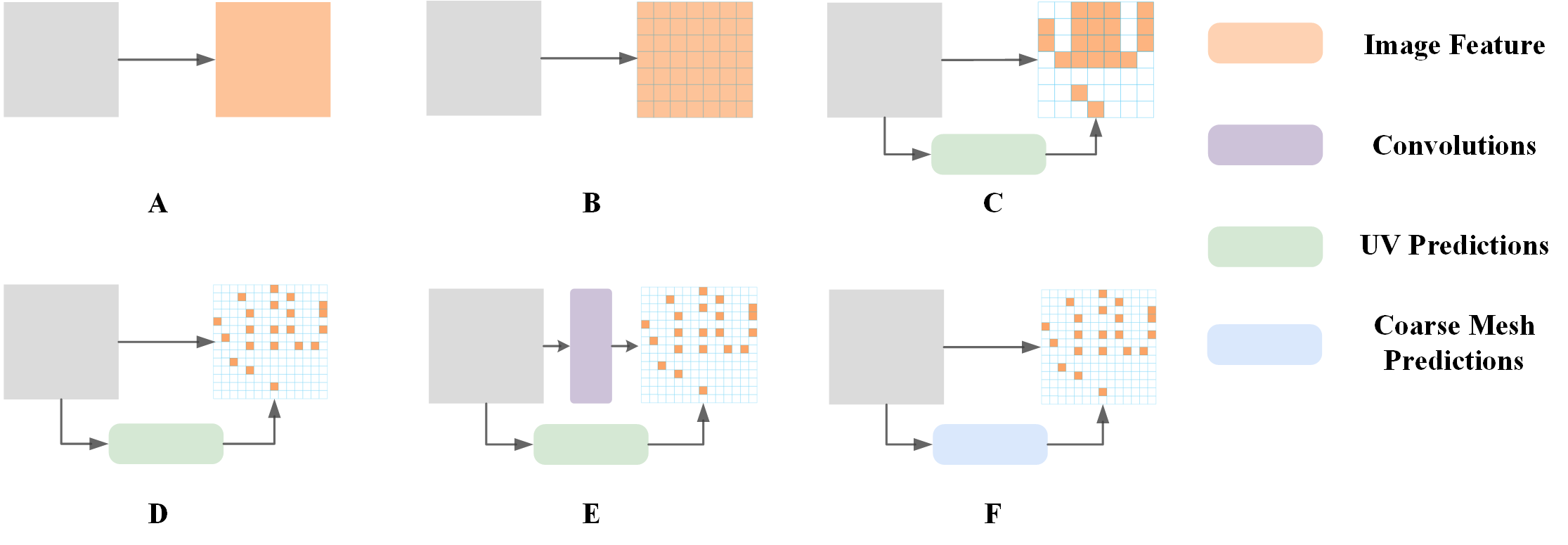}  
    \caption{An illustration demonstrates various designs of token generators. The grids colored in red represent the sampled points. a) Global feature; b) Grid sampling; c) Keypoint-guided sampling on the original feature map; d) Keypoint-guided sampling with 4x upsampling, resulting in an enhanced feature; e) Keypoint-guided sampling with 4x upsampling, where the feature is further improved by convolution; f) Coarse-mesh-guided point sampling with 4x upsampling.}  
    \label{fig:token_generator}  
\end{figure*}

\section{Related Work} 
In this section, we briefly review existing methods of hand mesh reconstruction which usually include two main components: a token generator and a mesh regressor. Token generator processes the backbone image feature and generates tokens fed to the decoder. Mesh regressor decodes the input tokens into 3D mesh directly or parametric hand model coefficient.

\subsection{Hand Mesh Reconstruction}
Estimating the 3D hand mesh from a single image has been widely researched. ~\cite{Zhang_Li_Mo_Zhang_Zheng_2019} proposes an end-to-end framework to recover hand mesh from a monocular RGB image. They use the 2D heatmap as input tokens and fully convolutional and fully connected layers to regress the MANO~\cite{Romero_Tzionas_Black_2017} parameters.

Transformer~\cite{Vaswani_Shazeer_Parmar_Uszkoreit_Jones_Gomez_Kaiser_Polosukhin_2017} has shown powerful performance in language and vision tasks which could model long range relation among input tokens. MetaFormer~\cite{Yu_Luo_Zhou_Si_Zhou_Wang_Feng_Yan_2022} argues that the general architecture of transformers instead of the specific token mixer is the key player. They replace the self-attention module with a simple spatial pooling operator and achieve competitive performance with fewer parameters and less computation. METRO~\cite{Lin_Wang_Liu_2021} extracts a single global image feature with a convolutional neural network and performs position encoding by repeatedly concatenating the image feature with 3D coordinates of a mesh template. A multi-layer transformer encoder with progressive dimensionality reduction regresses the 3D coordinates of mesh vertices with these input tokens. Due to the constraints of memory and computation, the transformer only processes a coarse mesh by sub-sampling twice with a sampling algorithm~\cite{Ranjan_Bolkart_Sanyal_Black_2018}, and Multi-Layer Perceptrons (MLPs) are then used to upsample the coarse mesh to the original mesh.

Graph convolutional neural network(GCNN)~\cite{Kipf_Welling_2016} is good at modeling the local interaction between neighbor vertices, thus it is very appropriate for mesh reconstruction. Pose2Mesh~\cite{Choi_Moon_Lee_2020} designs a cascaded architecture to regress 3D mesh vertices from 2D pose directly using GCNN. MeshGraphormer~\cite{Lin_Wang_Liu_2021} combines the ability of transformer and GCNN presenting a graph-convolution-reinforced transformer to model both local and global interactions.

Instead of extracting a global feature from the input image, pointHMR~\cite{kim2023sampling} argues that sampling features guided by vertex-relevant points could better utilize the correspondence between encoded features and spatial positions. They conduct feature sampling by element-wise multiplication of image feature and 2D heatmap trained by projection of 3D mesh vertices. These sampled features are then fed into the transformer encoder with progressive attention mask as the form of vertex token. The progressively decreased local connection range realized by constraining the attention mask encourage the model to consider the local relationship between neighbor vertices. They also use linear projection to reduce the dimension of the encoded token and upsampling algorithm~\cite{Ranjan_Bolkart_Sanyal_Black_2018} to expand the sparse vertices into original dense vertices.

\subsection{Lightweight Networks}
To achieve real-time hand mesh reconstruction, many lightweight networks have been studied for years. FastViT~\cite{Vasu_Gabriel_Zhu_Tuzel_Ranjan_2023} is a hybrid vision transformer architecture which obtains state-of-the-art latency-accuracy trade-off by structural reparameterization and train-time overparametrization techniques. MobRecon~\cite{Chen_Liu_Dong_Zhang_Ma_Xiong_Zhang_Guo_2022} designs multiple complicated modules to improve efficiency, including a stacked 2D encoding structure, a map-based position regression 2D-to-3D block and a graph operator based on spiral sampling~\cite{Lim_Dielen_Campen_Kobbelt_2019}. FastMETRO~\cite{Cho_Youwang_Oh} identifies the performance bottleneck of encoder-based transformers is caused by token design. They propose an encoder-decoder architecture to disentangle the interaction among input tokens which reduces the parameter.

%% file: sec_simhand/3_method.tex
\section{Method} 
In our research, we dissected the existing methods into two key components: a token generator and a mesh regressor. However, defining the optimal core structure for each of these modules remains a challenging task. For each module, we start with a fundamental, intuitive structure, and then progressively incorporate the most commonly used components, which we have abstracted from state-of-the-art (SOTA) methods.

Given that these two modules, the token generator and the mesh regressor, operate in tandem, it's important to keep one constant when analysing the other. In practical terms, we first conduct experiments on the mesh regressor while keeping the token generator, as depicted in Figure~\ref{fig:token_generator}-B, constant. Then, we apply the mesh regressor configuration that demonstrated the best performance to the token generator in subsequent experiments.



\subsection{Token Generator} 

Given a single image of dimensions $\{H,W\}$, our model utilizes a backbone to extract image features $X_{b}^{\in \frac{H}{32}\times \frac{W}{32} \times C}$. The token generator $T$ takes  $X_{b}$  as input and produces tokenized mesh feature $X_{m}^{\in N \times C} $, where $N$ denotes the number of sampled points. Thus, we can express this as $X_{m}=T(X_{b})$. 

Starting with the simplest implementation, we apply a single spatial pooling (Figure~\ref{fig:token_generator}-A). This approach establishes a surprisingly competitive baseline, comparable to the Fastmetro ~\cite{Cho_Youwang_Oh}. Changing spatial pooling to point sample (Figure~\ref{fig:token_generator}-B) improves the performance. To further improve the quality of the feature, we follow the MobRecon~\cite{Chen_Liu_Dong_Zhang_Ma_Xiong_Zhang_Guo_2022} model and conduct keypoint-guided point sampling(Figure~\ref{fig:token_generator}-C). However, this modification did not yield any noticeable improvement. 

Upon visual inspection, it appears that a $7\times7$ resolution is not sufficiently discriminating. Consequently, we apply  deconvolution on $X_{b}$ to sample the feature map to $14\times14$ then $28\times28$ (Figure~\ref{fig:token_generator}-D), respectively. This approach results in progressive improvement, but it does not work for $8 \times$ deconvolution or larger. 

Models such as MobRecon~\cite{Chen_Liu_Dong_Zhang_Ma_Xiong_Zhang_Guo_2022} and PointHMR~\cite{kim2023sampling} report improvements by enhancing features, for example, using a FPN-like structure or stacked blocks. In our study, we tested different $4 \times$ upsample schemes, including double $2 \times$ upsampling, directly $4 \times$ upsampling, and adding more convolution layers during the upsampling process(Figure~\ref{fig:token_generator}-E). Although these schemes vary in computational complexity, their performance remains consistent. 

We also tested the coarse mesh sampling method proposed by FastMETRO ~\cite{Cho_Youwang_Oh}. This method (Figure~\ref{fig:token_generator}-F) generates denser points compared to keypoint-guided sampling but does not offer any significant advantages. Detailed results are shown in table~\ref{tab:ablation_token_generator}. These experiments suggest that keypoint-guided point sampling at an appropriate resolution is a crucial structure for the token generator. As such, feature enhancement and exhaustive point sampling are not as necessary as initially thought.

\subsection{Mesh Regressor} 
\begin{figure}[t]
    \includegraphics[width=1.0\linewidth]{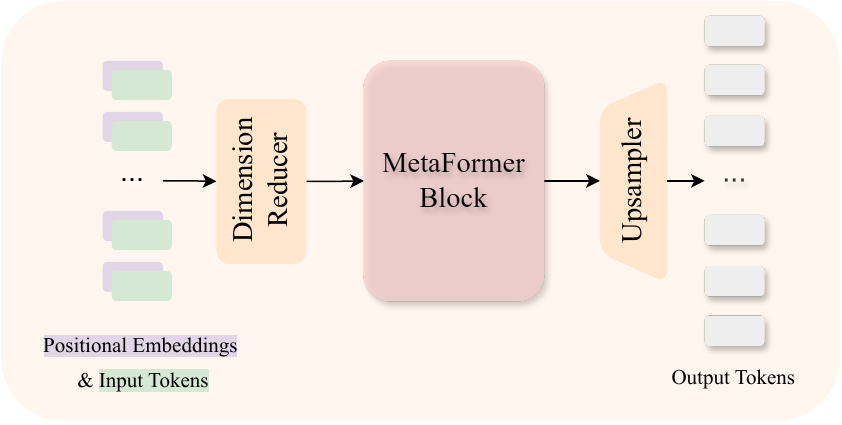} 
    \caption{Architecture of decoder layer in mesh regressor. It is composed of sequentially connected dimension reduce layer, metaformer block and upsample layer.}
    \label{fig:encoder_layer}
\end{figure}

The mesh regressor $R$ takes the tokenized mesh feature $X_{m}^{\in N \times C}$ as input and outputs predicted meshes Figure~\ref{fig:encoder_layer}.  ~\cite{Cho_Youwang_Oh}~\cite{Chen_Liu_Dong_Zhang_Ma_Xiong_Zhang_Guo_2022}  adopts a multi-stage approximation approach and proposes various methods to formulate the topology relationship between joints and mesh. Finding their intersection components, we construct a cascading upsampling mesh regressor $R$ using a series of decoder layers:
\begin{equation}
R = H_k H_{k-1}  ...  H_0
\end{equation}
Each decoder layer $H_k$ takes the calculated tokens $T_k$ as input, then subsequently processes these using a dimension reduce layer, metaformer, and upsample layer:
\begin{equation}
H_k(X_k) = U_k(MF_k(P_k(X_k)))
\end{equation}
where $U_k$, $P_k$ denotes the $k_{th}$ upsample layer and dimension reduce layer respectively, each composed of a single-layer MLP. The upsample layer increases token numbers, while the dimension reduce layer modifies channel shapes. $MF_k$ denotes the $k_{th}$ metaformer block,  $T_k$ is the $k_{th}$ output token where $X_{k+1}=H_k(X_k)$, $X_{0}=X_{m}$.

Let $d_k$ be the output dimension of $U_k$ and token numbers of $MF_k$, $n_k$,$c_k$ and $m_k$ are the block number, tokenmixer and block dimensions for $MF_k$. We start with the first layer $MF_0$ to demonstrate its operation. For the ${N \times C}$ shaped tensor $T_{0}$, $P_0$ projects it to ${N \times c}$, which is then processed by $MF_0$ and outputs a tensor of the same shape. Subsequently, $U_k$ upsamples it to ${d \times c}$. The following decoder layers repeat this procedure to output $X_{k}$.

 We began from a baseline where $\{k=1, n=\{1\}, d=\{778\}, m=\{identity\}\}$, which yielded competitive performance despite its simplicity. We then increase flops by enlarge $n$ but observe no improvement. Inspired by ~\cite{Chen_Liu_Dong_Zhang_Ma_Xiong_Zhang_Guo_2022}, We sequentially add blocks with an increasing value of $d$. When $k \leq 3$, Significant performance improvements are observed. However, as $k$ continues to increase beyond this point, no further gains are detected. Moreover, Modifying the token mixer from $ide$ to $attn$ also beneficial. However, for fixed $d$, simply increasing $n$ did not improve performance. According to our experimental findings, the core function of each decoder layer is to incrementally elevate the number of tokens from an initial quantity of 21 up to 778. Additional strategies like augmenting computational workload or altering intricate specifics of the network appear to have minimal impact.  In our best practice, parameters were set to  $\{k=3, n=\{1,1,1\}, d=\{21, 84, 336\}, m=\{attn,attn,attn\}\}$.

Existing hand joints and mesh topology modulation approaches stand out due to their ability to incorporate spatial information. However, their heuristic design is heavily reliant on hyperparameters and can be labor-intensive. Recognizing these strengths, we propose a novel method that modulates spatial relations without requiring manual design or additional computational resources. We achieve this by introducing learnable position embedding parameters $emb_k$ to each output tensor $X_k$ where 

\begin{equation}
X_k = X_k+emb_k
\end{equation}

\begin{figure*}[t!]
    \includegraphics[width=1.0\linewidth]{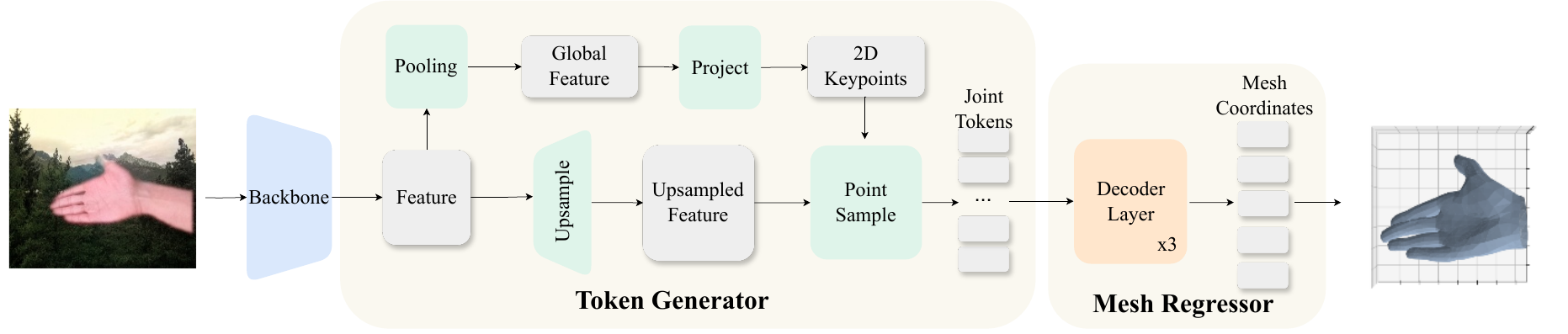} 
    \caption{ Overview of our architecture. The architecture of our model proceeds as follows: Firstly, the image feature $X_b$ is extracted via a backbone network. These features are then passed to our token generator module, responsible for predicting 2D keypoints and performing point sampling on the upsampled feature map, thus generating joint tokens. Next, these joint tokens are input into our mesh regressor module, which carries out the mesh prediction to get the final coordinates.}
    \label{fig:overview}
\end{figure*}

\subsection{Framework Design}
As discussed above, the image feature extracted by the backbone is sequentially processed by both the token generator and the mesh regressor. The overall framework can be simply computed by $R(T(X_{b}))$.

The core structures form the basis of the overall structure, which is depicted in Figure~\ref{fig:overview}. Given an input image of size $\{H, W\}$, we conduct point sampling guided by the predicted 21 keypoints at a resolution of $H/8, W/8$. For image classification style backbones like Fast-ViT, we apply a 4x upsample deconvolution to its final layer. However, for segmentation style backbones like HRNet, we directly use the feature on the corresponding resolution. In the mesh regressor, we apply position encoding before each MetaFormer block. Although this is not regarded as a core structure, it serves as a beneficial addition.

\subsection{Loss Functions}
The method proposed in this paper is trained with supervision for vertices, 3D joints, and 2D joints. In our implementation, both the 2D joints, denoted as $J_{2d}$, and the vertices, denoted as $V_{3d}$, are directly predicted by the model's output. The 3D joints, represented as $J_{3d}^{'}$, are calculated using the equation $J_{3d} = J \times V_{3d}$, where $J$ signifies the regression matrix. All of these components utilize the L1 loss to compute the discrepancy between the ground truth and the predictions. The losses for the vertex, 3D joint, and 2D joint, denoted as $L_{vert}$, $L_{J_{3d}}$, and $L_{J_{2d}}$, are respectively formulated as follows:

\begin{equation}
L_{J_{3d}} = \frac{1}{M_{J_{3d}}}\| J_{3d}-J_{3d}^{'}\|_1
\end{equation}

\begin{equation}
L_{J_{2d}} = \frac{1}{M_{J_{2d}}}\| J_{2d}-J_{2d}^{'}\|_1
\end{equation}

\begin{equation}
L_{vert} = \frac{1}{M_{V_{3d}}}\| V_{3d}-V_{3d}^{'}\|_1
\end{equation}

Here, $J_{3d} \in R^{M\times 3}$ represents all the ground truth points, and the symbols annotated with primes denote the predicted values. The overall loss function is defined as:

\begin{equation}
L = w_{3d}L_{J_{3d}} + w_{2d}L_{J_{2d}} + w_{vert}L_{vert}
\end{equation}

Given that the primary objective of this study is mesh prediction, 2D keypoints only affect point sampling and thus do not need to be highly accurate, we have accordingly adjusted the coefficients $w_{3d}$, $w_{2d}$, and $w_{vert}$ to 10, 1, and 10, respectively.

%% file: sec_simhand/4_experiment.tex
\section{Experiments}

\label{sec:exp}
\subsection{Implementation Details}

%

Our network is implemented based on Pytorch~\cite{paszke2017automatic}. We use HRNet64\cite{wang2020deep} and FastViT-MA36~\cite{Vasu_Gabriel_Zhu_Tuzel_Ranjan_2023} as our backbones, with their initial weights pre-trained on ImageNet. We use the AdamW~\cite{kingma2014adam} optimizer to train our network, with a total of 100 epochs. The learning rate is initially set to 5e-4, and then adjusted to 5e-5 after 50 epochs. We train the network with eight RTX2080Ti GPUs, with a batch size of 32 per GPU. It costs 7 hours training with FastViT-MA36 backone and 11 hours with HRNet. The features of intermediate layers are directly fed to the Token Generator without extra upsampling layer when the backbone is HRNetw64. The Mesh Regressor has three Encoder Layers, with the corresponding input token numbers being [21, 84, 336], output token numbers being [84, 336, 778], and feature dimensions being [256, 128, 64] respectively. We adopt Attention as the default token mixer, as its performance is slightly better.

\subsection{Datasets}

Our primary experiments and analyses are conducted on the FreiHAND~\cite{zimmermann2019freihand} dataset. In order to validate the generalization of our method, we also do experiments on the large-scale 3D hand-object dataset, DexYCB~\cite{chao2021dexycb}. The FreiHAND dataset contains 130,240 training samples and 3,960 testing samples. DexYCB contains 406,888 training samples and 78,768 testing samples.

\subsection{Evaluation Metrics}

\begin{table*}[t]
\centering
\resizebox{0.8\textwidth}{!}{
\begin{tabular}{c c | c c c c c} 
 \hline
 Method & Backbone & PA-MPJPE $\downarrow$ & PA-MPVPE $\downarrow$ & F@05 $\uparrow$ & F@15 $\uparrow$ & FPS \\
\hline
 I2L-MeshNet~\cite{moon2020i2l} & ResNet50 & 7.4 & 7.6 & 0.681 & 0.973 & 72 \\ 
 CMR ~\cite{chen2021camera} &  ResNet50 & 6.9 & 7.0 & 0.715 & 0.977 & - \\ 
 I2UV ~\cite{chen2021i2uv} & ResNet50 & 7.2 & 7.4 & 0.682 & 0.973 & -  \\ 
 Tang et al. ~\cite{tang2021towards} & ResNet50 & 6.7 & 6.7& 0.724 & 0.981 & 47 \\
 MobRecon ~\cite{Chen_Liu_Dong_Zhang_Ma_Xiong_Zhang_Guo_2022} &  DenseStack & 6.9 & 7.2 & 0.694 & 0.979 & 80 \\
METRO ~\cite{Zhang_Li_Mo_Zhang_Zheng_2019} & HRNet & 6.7 & 6.8 & 0.717 & 0.981 & 27 \\ 
MeshGraphomer~\cite{Lin_Wang_Liu_2021} &HRNet &6.3 & 6.5 & 0.738 & 0.983 & 24 \\
FastMETRO~\cite{Cho_Youwang_Oh} & HRNet & 6.5 & 7.1 & 0.687 & 0.983 & 28 \\ 
Deformer ~\cite{yoshiyasu2023deformable} & HRNet & 6.2 & 6.4 & 0.743 & 0.984 & - \\
PointHMR ~\cite{kim2023sampling} & HRNet & 6.1 & 6.6 & 0.720 & 0.984 &  - \\ 
 FastViT ~\cite{Vasu_Gabriel_Zhu_Tuzel_Ranjan_2023} & FastViT-MA36 & 6.6 & 6.7 & 0.722 & 0.981 & 84 \\
 \hline
 Ours & HRNet & \textbf{5.8} & \textbf{6.1} & \textbf{0.766} & \textbf{0.986} & 33 \\
 Ours & FastViT-MA36 &\textbf{5.7} & \textbf{6.0} & \textbf{0.772} & \textbf{0.986} & 70 \\

 \hline
\end{tabular}
}

\caption{Results on the FreiHAND dataset. Our results are shown in bold. ``-" indicates not reported. Our results surpass all existing methods in terms of accuracy metrics. }
\label{table:freihand_results}
\end{table*}

To evaluate the accuracy of 3D Hand Mesh Reconstruction methods, we adopt five metrics: Procrustes-aligned mean per joint position error (PA-MPJPE), Procrustes-aligned mean per vertex position error (PA-MPVPE), mean per joint position error (MPJPE), mean per vertex position error (MPVPE), and F-Score. PA-MPJPE and PA-MPVPE refer to the MPJPE and MPVPE after aligning the predicted hand results with the Ground Truth using Procrustes alignment, respectively. These two metrics do not consider the impact of global rotation and scale.

\subsection{Results}


\textbf{Comparison with previous methods} To validate our proposed modules., we adopted HRNet and FastViT-MA36 as backbones for non-real-time and real-time methods respectively, following established models ~\cite{Zhang_Li_Mo_Zhang_Zheng_2019}~\cite{Lin_Wang_Liu_2021}~\cite{Cho_Youwang_Oh}~\cite{Vasu_Gabriel_Zhu_Tuzel_Ranjan_2023}. For fair comparison, we provide performance metrics without Test-Time Augmentation (TTA) and FPS without TensorRT optimization. Table \ref{table:freihand_results} shows that our method, despite being slightly slower than FastViT, improves PA-MPJPE by 0.9mm. Compared to transformer-based methods, our approach demonstrates superior speed and performance, while only requiring 10\% of parameters, as shown in Table \ref{tab:compare_with_trans}.

\begin{figure*}[t]
    \begin{subfigure}{0.15\linewidth}
        \includegraphics[width=0.9\linewidth]{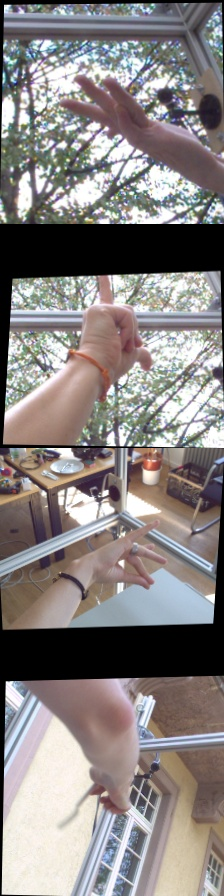} 
        \caption{Input}
        \label{fig:subim1}
    \end{subfigure}
    \begin{subfigure}{0.15\linewidth}
        \includegraphics[width=0.9\linewidth]{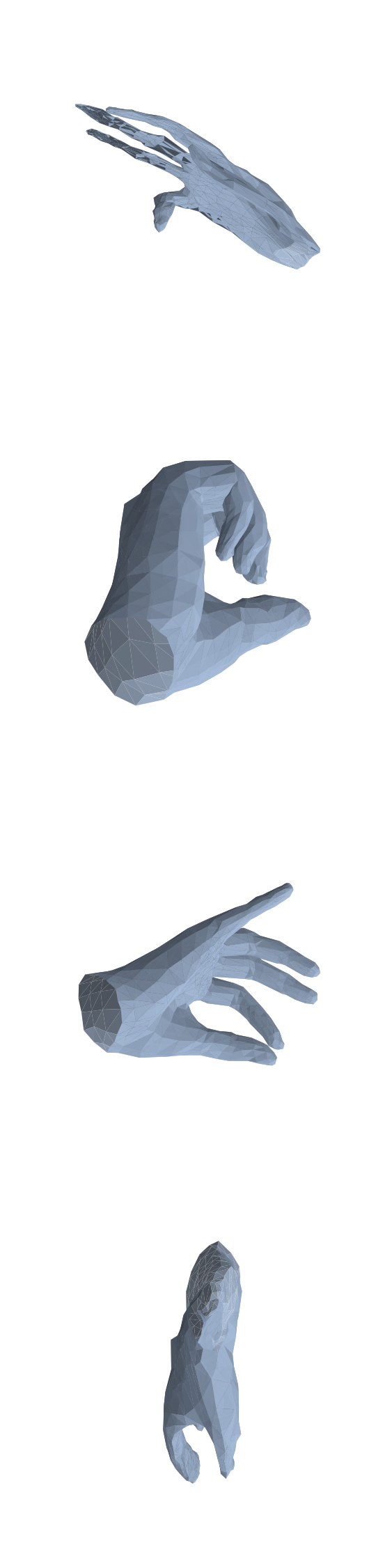}  
        \caption{MobRecon~\cite{Chen_Liu_Dong_Zhang_Ma_Xiong_Zhang_Guo_2022}}
        \label{fig:subim2}
    \end{subfigure}
    \begin{subfigure}{0.15\linewidth}
        \includegraphics[width=0.9\linewidth]{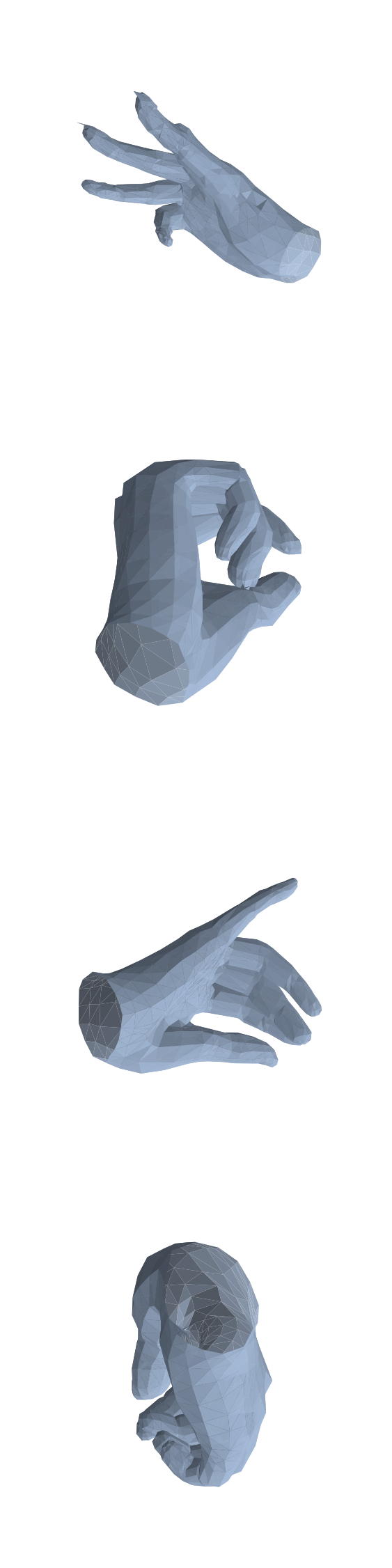}  
        \caption{METRO~\cite{Zhang_Li_Mo_Zhang_Zheng_2019}}
        \label{fig:subim3}
    \end{subfigure}
    \begin{subfigure}{0.15\linewidth}
        \includegraphics[width=0.9\linewidth]{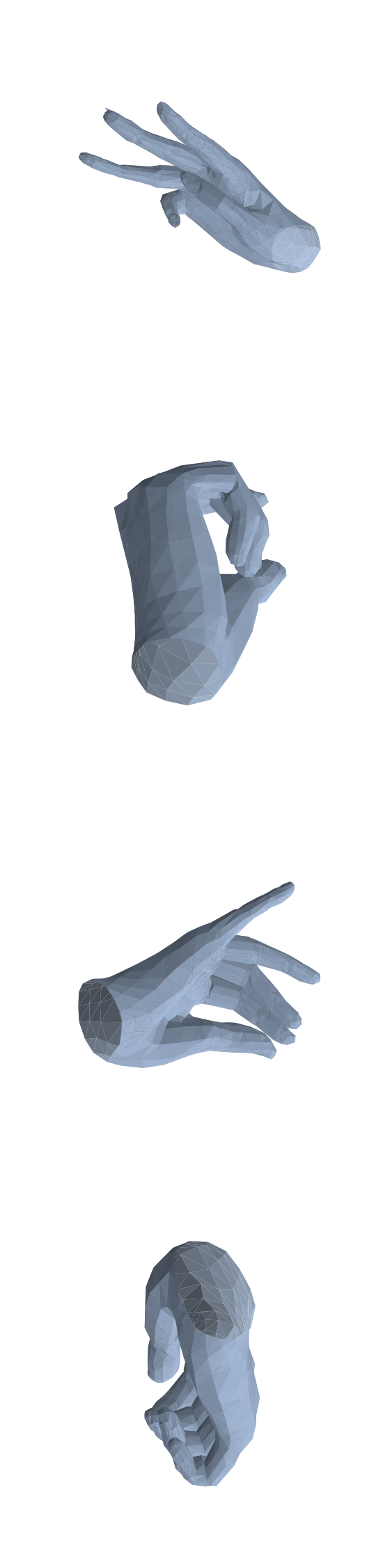}  
        \caption{MeshGraphormer~\cite{Lin_Wang_Liu_2021}}
        \label{fig:subim4}
    \end{subfigure}
    \begin{subfigure}{0.15\linewidth}
        \includegraphics[width=0.9\linewidth]{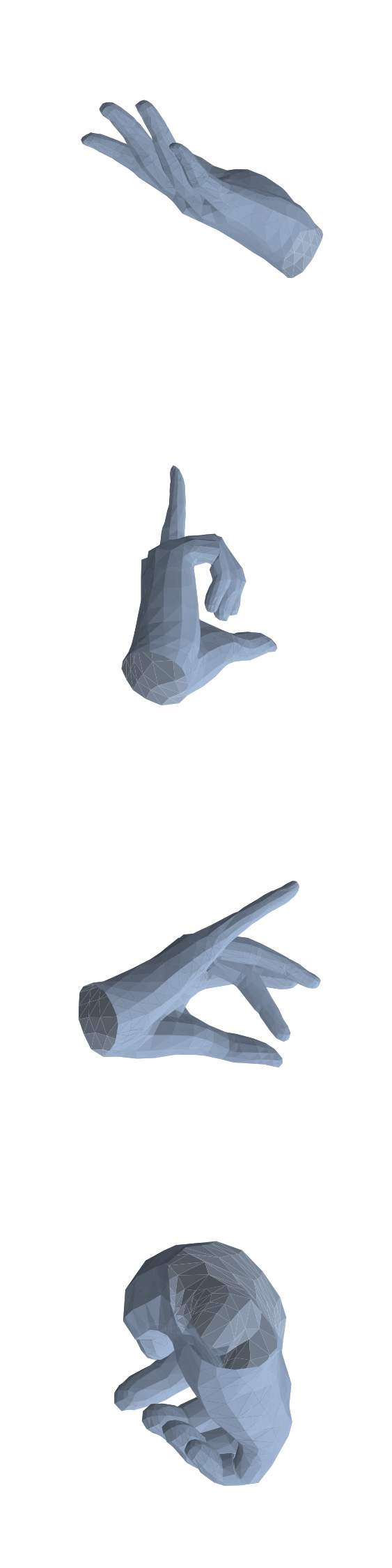}  
        \caption{Ours}
        \label{fig:subim5}
    \end{subfigure}
    \begin{subfigure}{0.15\linewidth}
        \includegraphics[width=0.9\linewidth]{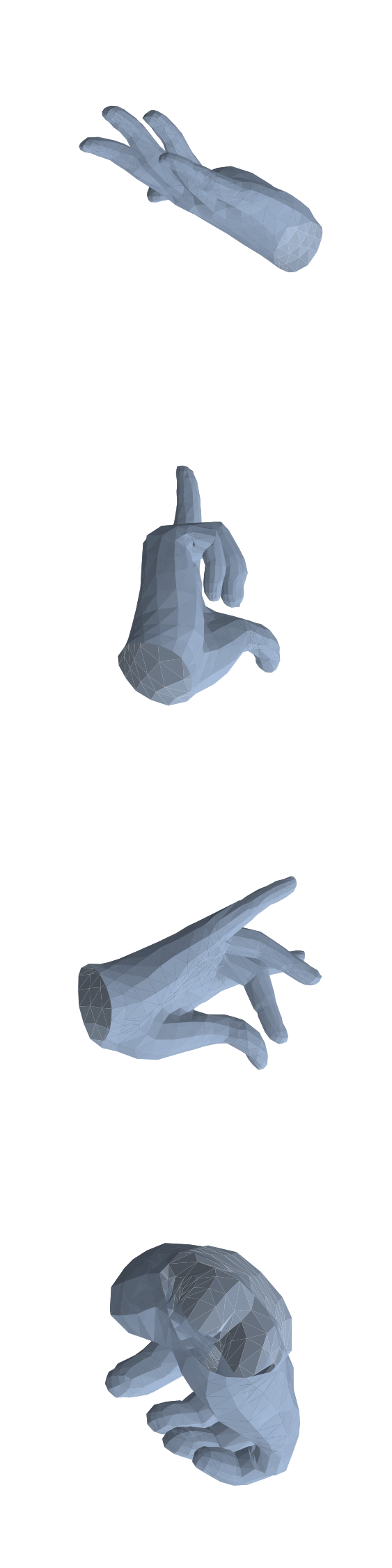}  
        \caption{GT}
        \label{fig:subim6}
    \end{subfigure}
\caption{Qualitative comparison between our method and other state-of-the-art approaches.}
\label{fig:Qualitative }
\end{figure*}

The qualitative comparison results are shown in the figure \ref{fig:Qualitative }. Compared to previous methods, our method produces more accurate hand reconstruction results.

\begin{table}[h]
    \centering
    \resizebox{\columnwidth}{!}{
    \begin{tabular}{c | c| c | c }
    \hline
    Method & $\#$Params & PA-MPJPE $\downarrow$ & PA-MPVPE $\downarrow$ \\
    \hline
    METRO ~\cite{Zhang_Li_Mo_Zhang_Zheng_2019} & 102M & 6.7 & 6.8 \\
    MeshGraphormer ~\cite{Lin_Wang_Liu_2021} & 98M & 6.3 & 6.5 \\
    FastMETRO ~\cite{Cho_Youwang_Oh}& 25M & 6.5 & 7.1 \\
    Ours & \textbf{1.9M} & \textbf{5.8} & \textbf{6.1} \\
    \hline
    \end{tabular}
    }
    \caption{\textbf{Comparison of transformer-based approaches.} $\#$Params refer to the network parameters that are not included within the backbone structure of the model. Our approach not only surpasses existing benchmarks in key metrics but also achieves a parameter reduction of one to two orders of magnitude.}
    \label{tab:compare_with_trans}
\end{table}

\textbf{Evaluation on DexYCB} We employed the large-scale hand-object dataset  DexYCB to validate our method's effectiveness and generalizability. As shown in Table \ref{tab:dexycb}, our model outperforms existing single-image input methods on all metrics. Significantly, we surpassed previous benchmarks by 1.5mm and 0.8mm on the MPJPE and MPVPE measures respectively, thereby setting new standards and demonstrating our method's broad applicability.

\captionsetup[subfigure]{labelformat=empty}

\begin{table}[h]
    \centering
    \resizebox{\columnwidth}{!}{
        \begin{tabular}{c| c c c c }
        \hline
        Method & PA-MPJPE $\downarrow$ & PA-MPVPE $\downarrow$ & MPJPE $\downarrow$ & MPVPE $\downarrow$  \\
        \hline
        METRO  ~\cite{Zhang_Li_Mo_Zhang_Zheng_2019} & 7.0 & - & - & - \\
        Spurr et al. ~\cite{spurr2020weakly}  & 6.8 & - & - & -  \\
        Liu et al.~\cite{liu2021semi}  & 6.6 & - & - & -   \\
        HandOccNet ~\cite{park2022handoccnet} & 5.8 & 5.5 & 14.0 & 13.1   \\
        MobRecon ~\cite{Chen_Liu_Dong_Zhang_Ma_Xiong_Zhang_Guo_2022} & 6.4  & 5.6 & 14.2  & 13.1   \\
        H2ONet ~\cite{xu2023h2onet} & 5.7  & 5.5  & 14.0  & 13.0   \\
        Ours & \textbf{5.5} & \textbf{5.5} & \textbf{12.4}  & \textbf{12.1}  \\ 
        \hline
        \end{tabular}
    }
    \caption{\textbf{Results on DexYCB}. Our method shows advantages on Procrustes-Aligned metrics and surpassed the previous methods by a large margin on non-Procrustes-Aligned metrics.}
    \label{tab:dexycb}
\end{table}

\subsection{Ablation Study}

To thoroughly validate the various parameter combinations, a large number of ablation experiments were conducted. For efficiency, all ablation experiments were implemented on smaller models (e.g., Hiera-tiny). After identifying the optimal parameter combination, it is then applied to the standard models to facilitate a fair comparison with existing methods.

The state-of-the-art backbone, Hiera-Tiny~\cite{ryali2023hiera}, is utilized in our study as a strong baseline. We conduct a series of ablation experiments on the FreiHAND dataset with the aim of examining the efficacy of the structure we propose.

\textbf{Effectiveness of Our Token Generator and Mesh Regressor}. In order to evaluate the effectiveness of our Token Generator and Mesh Regressor, we initially set up a standard baseline model. This model's Token Generator is constructed based on global features, while its Mesh Regressor is designed as a Multilayer Perceptron (MLP). We subsequently substitute these components with our proposed structures individually. The results of these experiments, detailed in Table \ref{tab:ablation_overall_module}, confirm that both modules, when incorporated in place of the original structures, contribute positively towards enhancing overall performance. When implemented together, these modifications lead to even further improvements.
\begin{table}[h]
    \centering
    \resizebox{0.8\columnwidth}{!}{
    \begin{tabular}{ c | c | c }
    \hline
    Method  & PA-MPJPE $\downarrow$ & PA-MPVPE $\downarrow$ \\
    \hline 
    Simple Baseline & 6.9 & 7.2 \\
    + mesh regressor &  6.5  & 6.8\\
    + token generator & 6.6  & 7.1\\
    + both  & 6.2  & 6.5 \\
    \hline
    \end{tabular}
    }
    \caption{\textbf{Ablation study of our proposed modules}. Each of these methods brings about enhancements when utilized individually. However, when these strategies are integrated, they yield an even more substantial improvement.}
    \label{tab:ablation_overall_module}
\end{table}

\begin{figure}[t]
    \begin{subfigure}{0.19\linewidth}
        \includegraphics[width=0.9\linewidth]{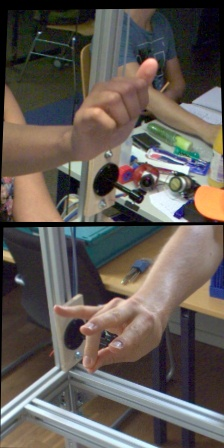} 
        \caption{Input}
        \label{fig:subim11}
    \end{subfigure}
    \begin{subfigure}{0.19\linewidth}
        \includegraphics[width=0.9\linewidth]{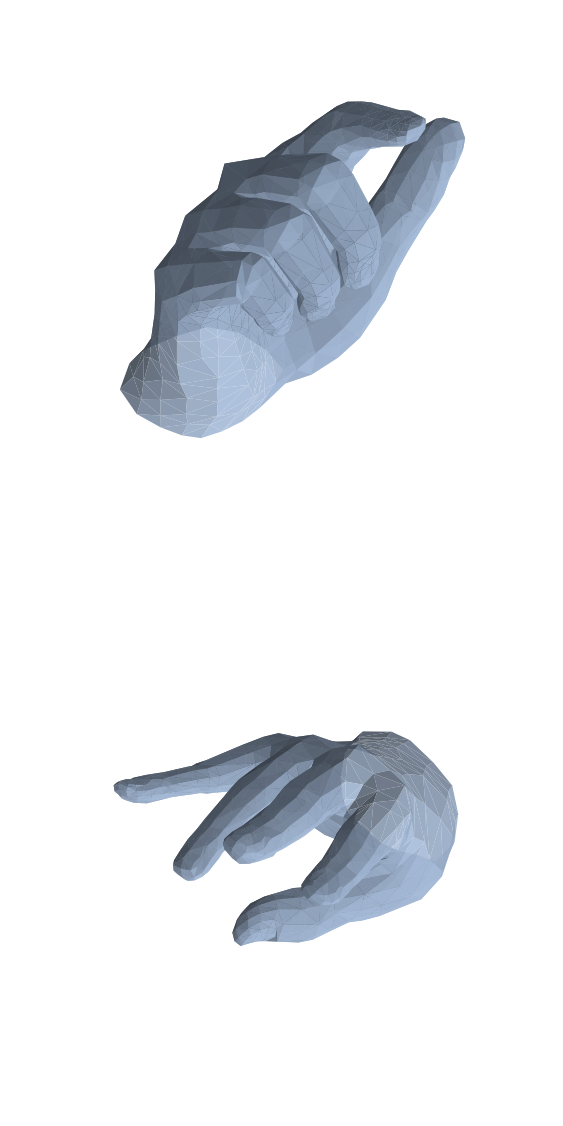}  
        \caption{global}
        \label{fig:subim21}
    \end{subfigure}
    \begin{subfigure}{0.19\linewidth}
        \includegraphics[width=0.9\linewidth]{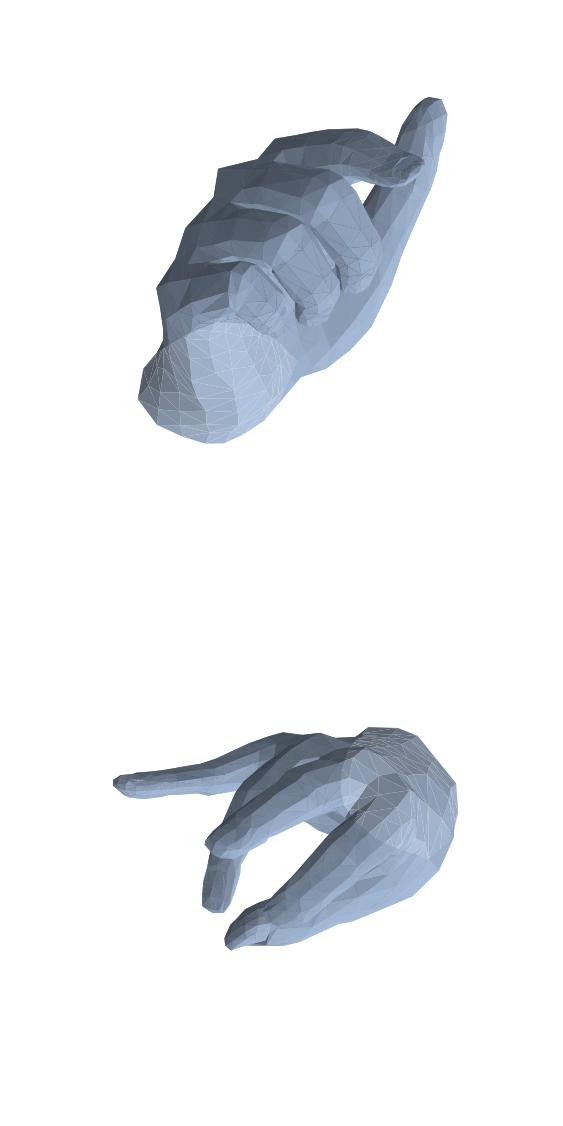}  
        \caption{coarse}
        \label{fig:subim31}
    \end{subfigure}
    \begin{subfigure}{0.19\linewidth}
        \includegraphics[width=0.9\linewidth]{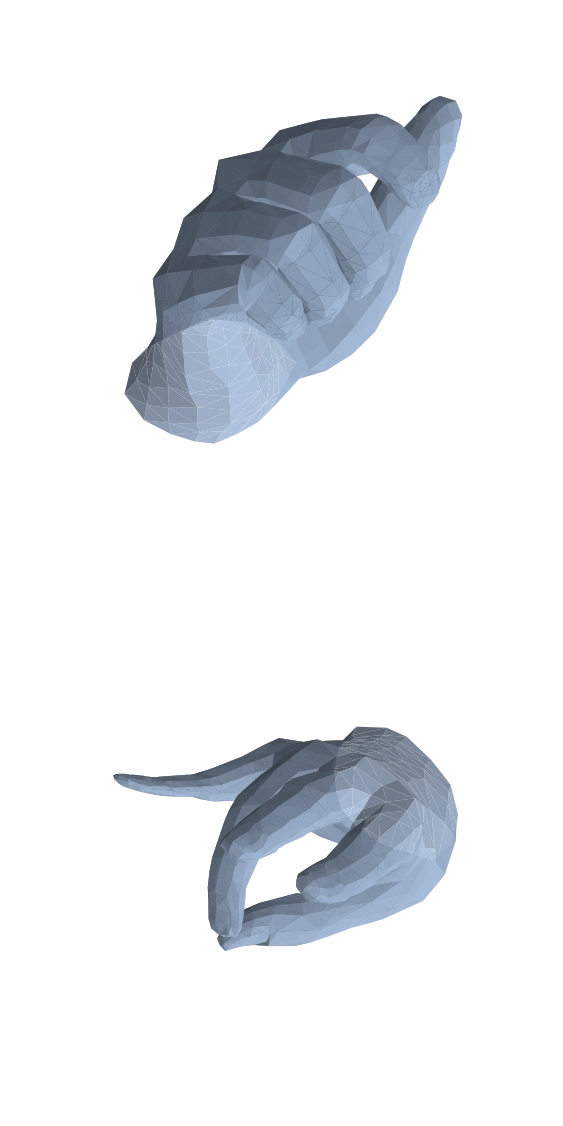}  
        \caption{upsampled}
        \label{fig:subim41}
    \end{subfigure}
    \begin{subfigure}{0.19\linewidth}
        \includegraphics[width=0.9\linewidth]{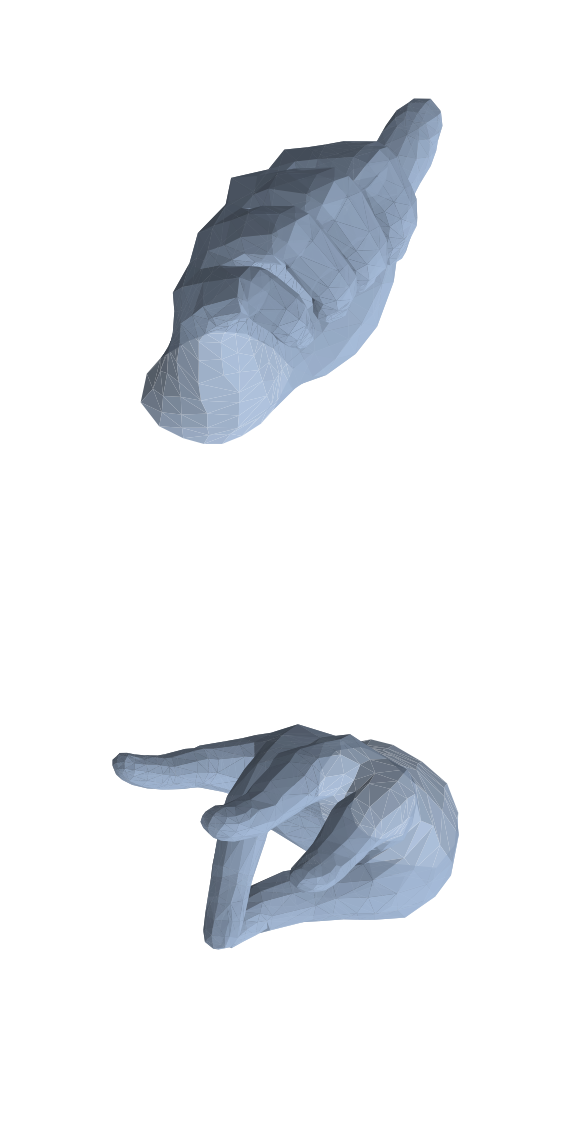}  
        \caption{GT}
        \label{fig:subim61}
    \end{subfigure}
\caption{{\bf Qualitative Comparison of Different Point Sampling Strategies.} The global/coarse feature fails in scenarios with detailed finger interactions, where upsampled feature works well.}
\label{fig:token_generator_results}
\end{figure}

\textbf{Analysing Core Structure of Token Generator.}
As shown in Table \ref{tab:ablation_token_generator}, performance using only global features is competitive. The grid sampling and point sampling on a $7 \times 7$ feature map show similar efficiencies. Increasing the resolution of the feature map to $28 \times 28$ through a single four-fold deconvolution improves performance. However, further optimization is not achieved  by replacing  single four-fold deconvolution with two layers of two-fold deconvolutions or adding more convolutions. Similarly, no improvement is observed when changing from point sampling to coarse mesh sampling. Qualitative comparison of different point sampling strategies is shown in \cref{fig:token_generator_results}

\begin{table}[h]
    \centering
     \resizebox{\columnwidth}{!}{
    \begin{tabular}{c | c | c | c}
    \hline
      Sample Method & Resolution & PA-MPJPE $\downarrow$ & PA-MPVPE $\downarrow$ \\ 
     \hline
     Global & 1x1 &6.5 & 6.8 \\
     Grid & 7x7 &6.3 & 6.6 \\
     Point & 7x7 & 6.3 &  6.6 \\
     Point &14x14 & 6.3 &  6.6 \\
     Point & 28x28 & 6.2 & 6.5 \\
     Point & 28 x28 enhanced& 6.2 & 6.5 \\
     Coarse mesh & 28x28 & 6.2 & 6.5 \\
     \hline
    \end{tabular}
    }
    \caption{\textbf{Ablation study of Our Token generator} A point sample at a resolution of 28x28 achieves optimal efficiency. Contrarily, increasing the number of sampled points or incorporating additional convolutional layers do not lead to any further improvements.}
    \label{tab:ablation_token_generator}
\end{table}

\begin{figure}[t]
    \begin{subfigure}{0.19\linewidth}
        \includegraphics[width=0.9\linewidth]{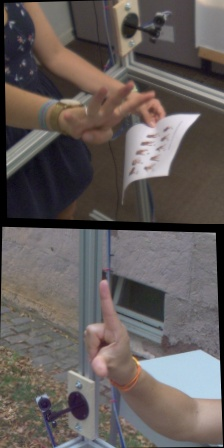} 
        \caption{Input}
        \label{fig:subim12}
    \end{subfigure}
    \begin{subfigure}{0.19\linewidth}
        \includegraphics[width=0.9\linewidth]{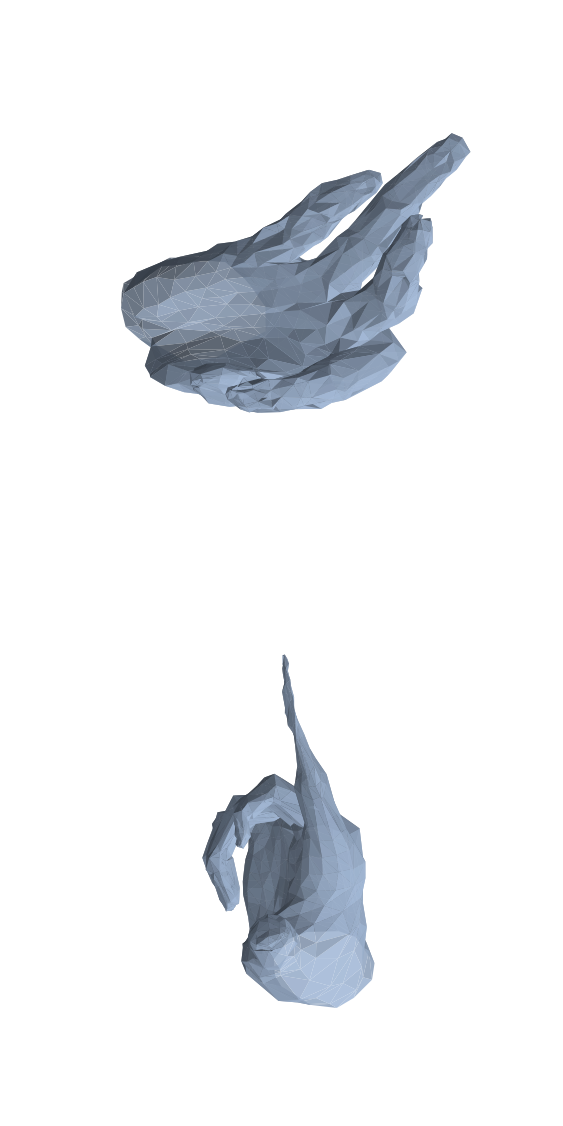}  
        \caption{one layer}
        \label{fig:subim22}
    \end{subfigure}
    \begin{subfigure}{0.19\linewidth}
        \includegraphics[width=0.9\linewidth]{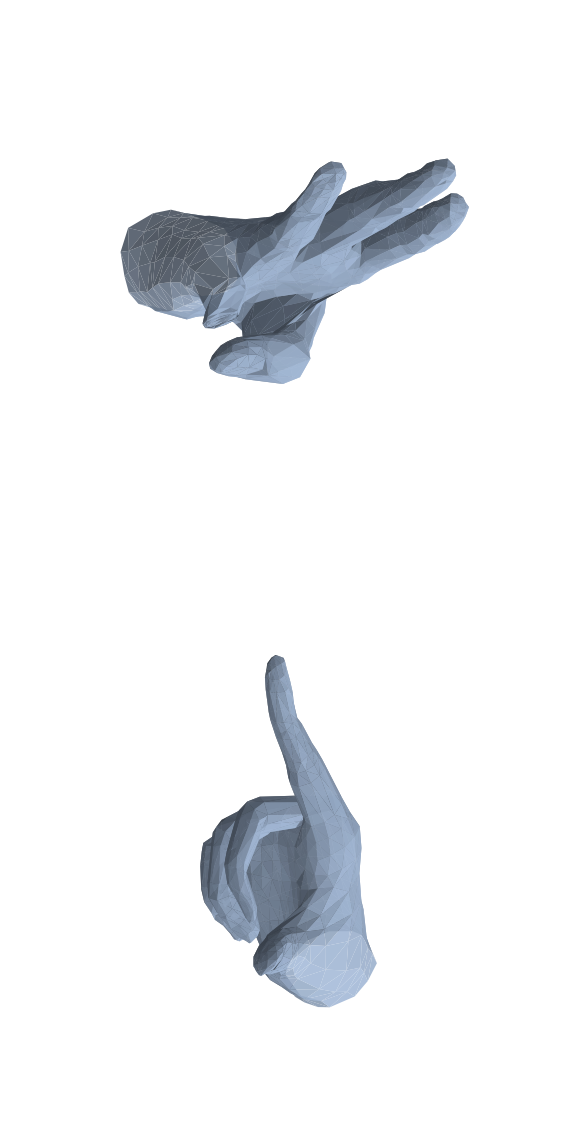}  
        \caption{two layer}
        \label{fig:subim32}
    \end{subfigure}
    \begin{subfigure}{0.19\linewidth}
        \includegraphics[width=0.9\linewidth]{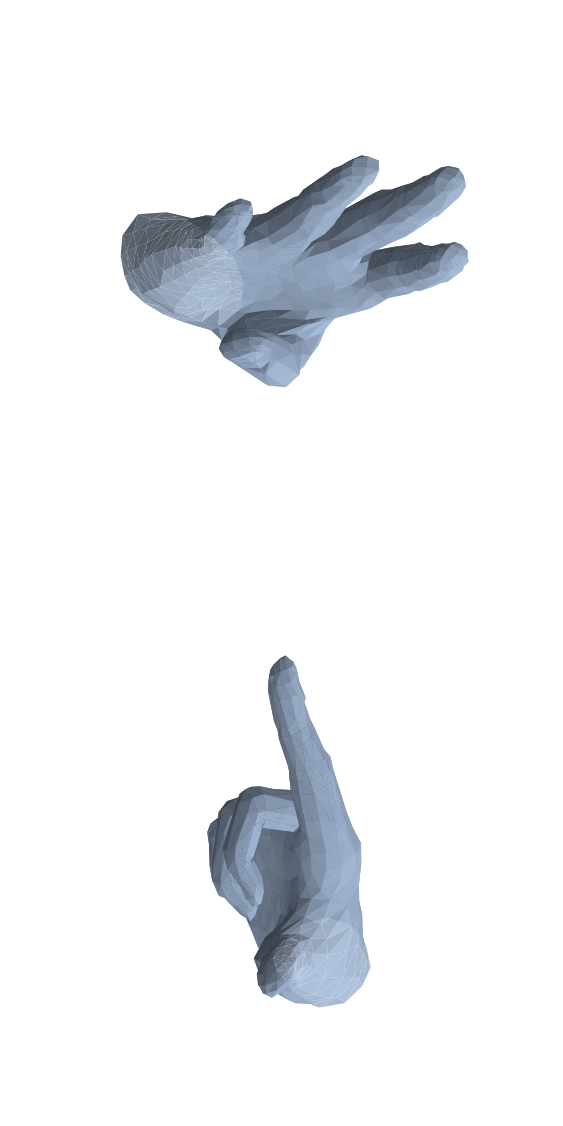}  
        \caption{three layer}
        \label{fig:subim42}
    \end{subfigure}
    \begin{subfigure}{0.19\linewidth}
        \includegraphics[width=0.9\linewidth]{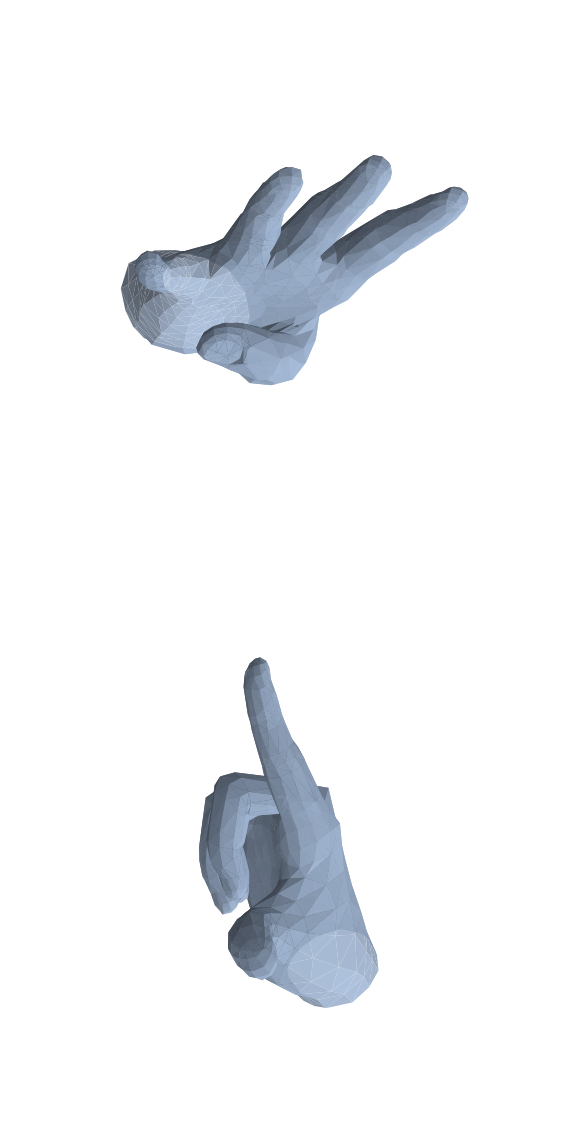}  
        \caption{GT}
        \label{fig:subim62}
    \end{subfigure}
%
\caption{{\bf Qualitative Comparison of the Number of Layers of Mesh Decoder.} When constrained to one, the reconstructed mesh tends to corrupt into unnatural shapes. Performance improves as the number of layers increases.}
\label{fig:mesh_decoder}
\end{figure}

\textbf{Analysing the Core Structure of the Mesh Regressor}. 
As shown in Table \ref{tab:ablation_of_layer_and_tokne_num}, for a single encoder layer, adding an extra encoder layer with a larger token number sharply increases performance by 0.3mm. The optimal setting consists of three encoder layers, with token numbers progressively multiplied by 4. As the layer number increases further, the marginal benefit becomes inconsequential and sometimes even decreases. Furthermore, as shown in Table \ref{tab:ablation_of_block_dims}, given a fixed set of token numbers, increasing computational complexity produces negligible differences in either block numbers or block dimensions in the encoder layer. A middle-sized block dimensions setting is optimal. Qualitative comparison of different upsample layers is shown in Figure \ref{fig:mesh_decoder}.

\begin{table}[h]
    \centering
    \resizebox{\columnwidth}{!}{
    \begin{tabular}{c |c | c | c}
    \hline
    Layer Nums & Token Nums & PA-MPJPE $\downarrow$ & PA-MPVPE $\downarrow$ \\
    \hline
        1 & [21] & 6.6 & 7.1  \\
        2 & [21, 256] & 6.3 & 6.6   \\
        2 & [21, 384] & 6.3 & 6.6  \\
        3 & [21, 256, 384] & 6.2 & 6.5  \\
        3 & \textbf{[21, 84, 336]} & 6.2 & 6.5  \\
        4 & [21, 128, 256, 384] & 6.2 & 6.5  \\
        4 & [21, 63, 126, 252] & 6.3 & 6.6  \\
    \hline
    \end{tabular}
    }
  \caption{\textbf{The Number of Upsampling Layers and Corresponding Token Numbers in Encoding Layers.} Three encoding layers yield optimal efficiency.}
    \label{tab:ablation_of_layer_and_tokne_num}
\end{table}

\begin{table}[h]
    \centering
    \resizebox{0.8\columnwidth}{!}{
    \begin{tabular}{c|c|c}
    \hline
    Dimensions & PA-MPJPE $\downarrow$ & PA-MPVPE $\downarrow$ \\
    \hline
    64, 32, 16 & 6.5 & 6.9 \\
    128, 64, 32 & 6.3 & 6.6 \\
    \textbf{256, 128, 64} & 6.2 & 6.5 \\
    512, 256, 12 & 6.2 & 6.5 \\
    1024, 512, 256 & 6.4 & 6.7 \\
    \hline
    Block Nums & PA-MPJPE $\downarrow$ & PA-MPVPE $\downarrow$ \\
    \hline
    \textbf{1, 1, 1} & 6.2 & 6.5 \\
    2, 2, 2 & 6.2 & 6.6 \\
    3, 3, 3 & 6.3 & 6.6 \\
    \hline
    \end{tabular}
    }
    \caption{\textbf{Dimensions and block nums}. Single layer blocks with middle sized dimensions are optimal.}
    \label{tab:ablation_of_block_dims}
\end{table}

\textbf{Position Encoding and Attention Mixer}. We utilized position encoding layer and attention mixer during ablation experiments because they are intuitively helpful. Based on our SOTA result, we remove position encoding layers to observe a slightly 0.1mm degrade. Similar thing happens when we substitute attention mixer to identity mixer, see \cref{tab:ablation_of_block_nums}.

\begin{table}[h]
    \centering
    \resizebox{0.8\columnwidth}{!}{
    \begin{tabular}{c|c|c}
    \hline
    Method & PA-MPJPE $\downarrow$ & PA-MPVPE $\downarrow$ \\
    \hline
    \textbf{hiera-tiny sota} & 6.2 & 6.5 \\
    
    Identity mixer& 6.3 & 6.6 \\
    w/o position emb & 6.3 & 6.6 \\
    \hline
    \end{tabular}
    }
    \caption{\textbf{Token mixer and position embedding}. When substitute attention mixer to identity mixer, or remove position encoding layer, the performance dropped sligntly by 0.1mm.}
    \label{tab:ablation_of_block_nums}
\end{table}

\begin{figure}[t]
    \begin{subfigure}{0.15\linewidth}
        \includegraphics[width=0.9\linewidth]{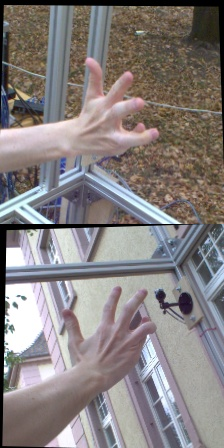} 
        \caption{Input}
        \label{fig:subim13}
    \end{subfigure}
    \begin{subfigure}{0.15\linewidth}
        \includegraphics[width=0.9\linewidth]{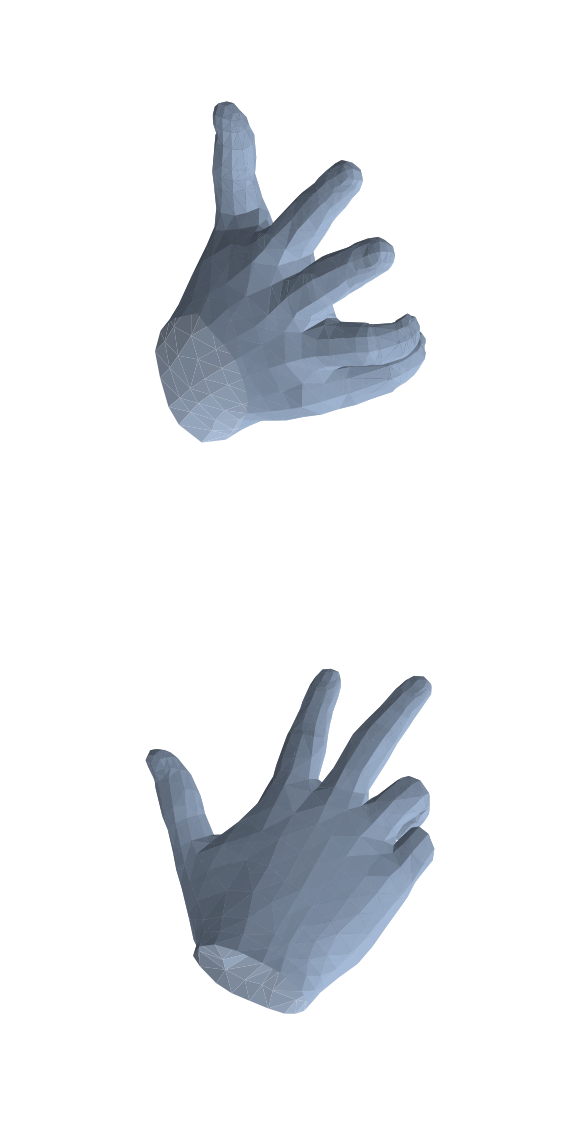}  
        \caption{ours}
        \label{fig:subim23}
    \end{subfigure}
    \begin{subfigure}{0.15\linewidth}
        \includegraphics[width=0.9\linewidth]{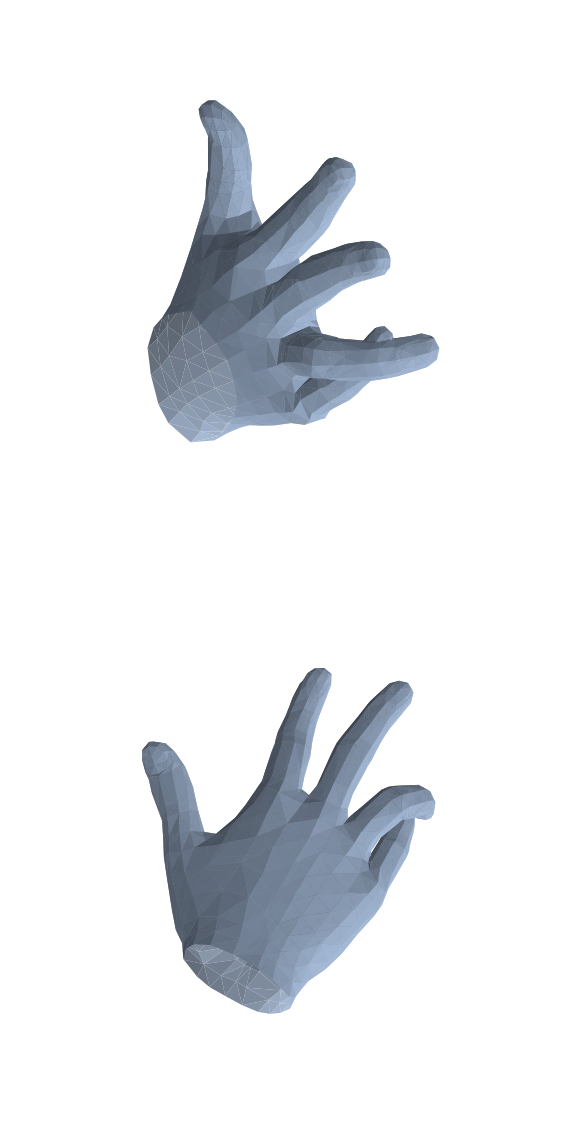}  
        \caption{gt}
        \label{fig:subim33}
    \end{subfigure}
    \begin{subfigure}{0.15\linewidth}
        \includegraphics[width=0.9\linewidth]{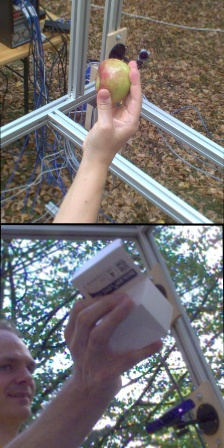} 
        \caption{Input}
        \label{fig:subim131}
    \end{subfigure}
    \begin{subfigure}{0.15\linewidth}
        \includegraphics[width=0.9\linewidth]{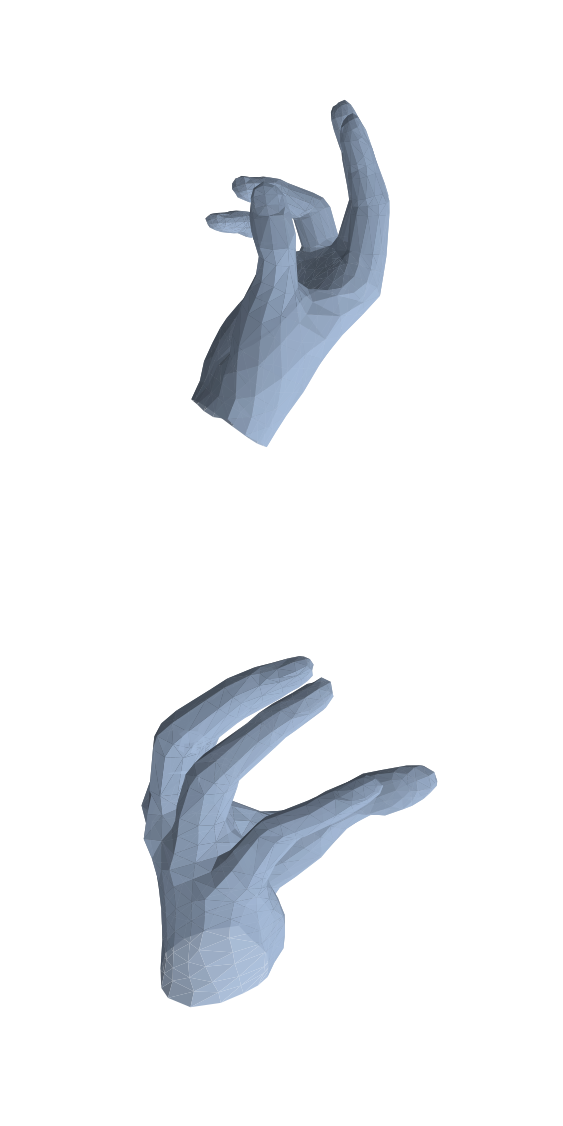}  
        \caption{ours}
        \label{fig:subim231}
    \end{subfigure}
    \begin{subfigure}{0.15\linewidth}
        \includegraphics[width=0.9\linewidth]{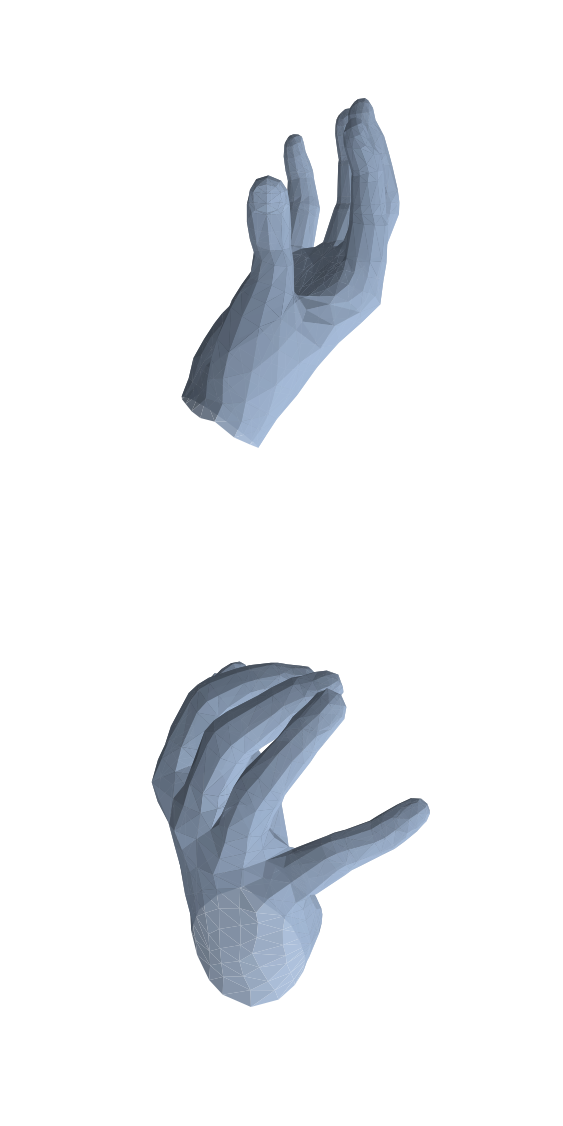}  
        \caption{gt}
        \label{fig:subim331}
    \end{subfigure}
\caption{{\bf Typical Failure Cases}. Failure cases are concentrated in scenes with self-occlusion and object occlusion. Some are difficult to discern due to the small area of exposure, while others present ambiguities caused by the occlusion.}
\label{fig:fail_cases}
\end{figure}

\textbf{Limits and Failure cases}. As mentioned earlier, our work is dedicated to summarizing and abstracting from existing work. Since no targeted optimization was performed, some failure cases  present in previous work remains challenging. These cases are concentrated in scenes with self-occlusion and object occlusion, see 
 \cref{fig:fail_cases}.

\captionsetup[subfigure]{labelformat=empty}

\captionsetup[subfigure]{labelformat=empty}

\captionsetup[subfigure]{labelformat=empty}

%% file: sec_simhand/5_conclution.tex
\section{Conclusion and Future Work} 

We observed shared advantages and disadvantages of typical structures. Based on these observations, we introduce the concept of the \textbf{core structure}. Through experiments, we revealed that a framework with the core structure could achieve high performance with limited computational load. We evaluated our approach quantitatively and qualitatively to demonstrate its effectiveness.

 However, our method is explicitly designed to reconstruct single hand gestures. Other scenarios, such as extreme lighting, occlusion, interactions, or out-of-distribution cases, showed no improvement over existing methods. Such cases require specifically designed methods.